\documentclass[journal]{IEEEtran}
\ifCLASSOPTIONcompsoc
  \usepackage[nocompress]{cite}
\else
  \usepackage{cite}
\fi
\usepackage{amsmath,amssymb,amsfonts}
\ifCLASSINFOpdf
   \usepackage[pdftex]{graphicx}
   \graphicspath{{../pdf/}{../jpeg/}}
   \DeclareGraphicsExtensions{.pdf,.jpeg,.png}
\else
   \usepackage[dvips]{graphicx}
   \graphicspath{{../eps/}}
   \DeclareGraphicsExtensions{.eps}
\fi
\usepackage{acronym}
\usepackage{multirow}
\usepackage{lscape}
\usepackage{longtable}
\usepackage{gensymb}
\usepackage{booktabs}
\usepackage[table,xcdraw]{xcolor}

\usepackage{supertabular,booktabs}

\usepackage{algorithmic}
\usepackage{array}
\usepackage{mdwmath}
\usepackage{mdwtab}
\usepackage{eqparbox}
\ifCLASSOPTIONcompsoc
  \usepackage[caption=false,font=footnotesize,labelfont=sf,textfont=sf]{subfig}
\else
  \usepackage[caption=false,font=footnotesize]{subfig}
\fi
\usepackage{fixltx2e}
\usepackage{stfloats}
\usepackage{url}
\ifCLASSINFOpdf
  \usepackage[pdftex]{thumbpdf}
\else
  \usepackage[dvips]{thumbpdf}
\fi
\begin{document}
\title{Approaches, Challenges, and Applications for Deep Visual Odometry: Toward to Complicated and Emerging Areas}
\author{Ke~Wang,
            ~Sai~Ma,
            ~Junlan~Chen,
            ~Fan Ren,            
%            ~Jianbo~Lu,~\IEEEmembership{Fellow,~IEEE} 
\thanks{K. Wang was with the School of Automobile Engineering, Chongqing University, China, 400044, also with the Key Lab of Mechanical Transmission, Chongqing University, China, 400044 (e-mail: kw@cqu.edu.cn).}% <-this % stops a space
\thanks{S. Ma was with the School of Automobile Engineering, Chongqing University, China, 400044,Chongqing University, China, 400044 (e-mail: masai@cqu.edu.cn).}% <-this % stops a space
\thanks{J. Chen is with shool of Economics and Management, Chongqing Normal University, Chongqing 401331, China (e-mail: junlanchen@cqnu.edu.cn).}% <-this % stops a space
\thanks{F. Ren is with Intelligent Vehicle R\&D Institute, Changan Auto Company, Chongqing 401120, China (e-mail: renfan@changan.com.cn).}% <-this % stops a space
%\thanks{J. Lu is with Research and Advanced Engineering, Ford Motor Company, Dearborn, MI 48121 USA (e-mail: jlu10@ford.com).}% <-this % stops a space
\thanks{Manuscript received April 19, 2005; revised August 26, 2015. (Corresponding author: Ke wang and Junlan Chen)}}


\markboth{Journal of \LaTeX\ Class Files,~Vol.~14, No.~8, August~2015}%
{Shell \MakeLowercase{\textit{et al.}}: Bare Demo of IEEEtran.cls for IEEE Journals}



\maketitle

\begin{abstract}
Visual odometry (VO) is a prevalent way to deal with the relative localization problem, which is becoming increasingly mature and accurate, but it tends to be fragile under challenging environments. Comparing with classical geometry-based methods, deep learning-based methods can automatically learn effective and robust representations, such as depth, optical flow, feature, ego-motion, etc., from data without explicit computation. Nevertheless, there still lacks a thorough review of the recent advances of deep learning-based VO (Deep VO). 
Therefore, this paper aims to gain a deep insight on how deep learning can profit and optimize the VO systems. We first screen out a number of qualifications including accuracy, efficiency, scalability, dynamicity, practicability, and extensibility, and employ them as the criteria. Then, using the offered criteria as the uniform measurements, we detailedly evaluate and discuss how deep learning improves the performance of VO from the aspects of depth estimation, feature extraction and matching, pose estimation. We also summarize the complicated and emerging areas of Deep VO, such as mobile robots, medical robots, augmented reality and virtual reality, etc. Through the literature decomposition, analysis, and comparison, we finally put forward a number of open issues and raise some future research directions in this field.
\end{abstract}


% \fi
\IEEEpeerreviewmaketitle

\section{Introduction}
\IEEEPARstart{V}{isual}  odometry is the problem of estimating the camera pose from consecutive images and is a fundamental capability required in many computer vision and robotics applications, such as Autonomous / Medical Robots, Augmented / Mixed / Virtual Reality and other complicated and emerging applications based on localization information, such as indoor and outdoor navigation, scene understanding, space exploration\cite{RN681,RN713,RN712}. 

Over the past decades, we have seen impressive progress on the classical geometry-based visual odometry. Superior performance on accurate and effective are demonstrated in both feature-based methods\cite{RN632, RN598, RN623} and direct methods\cite{RN637, RN636, RN635}, which pushes the classical geometry-based visual odometry towards real-world applications\cite{RN685, RN699, RN693,RN716,RN714}. However, the robustness of these methods cannot meet the requirements of high-reliability applications under challenging environments, such as illumination changing and dynamic environments, etc. 

\begin{figure*}[!t]
\centering
\includegraphics[width=6.5in]{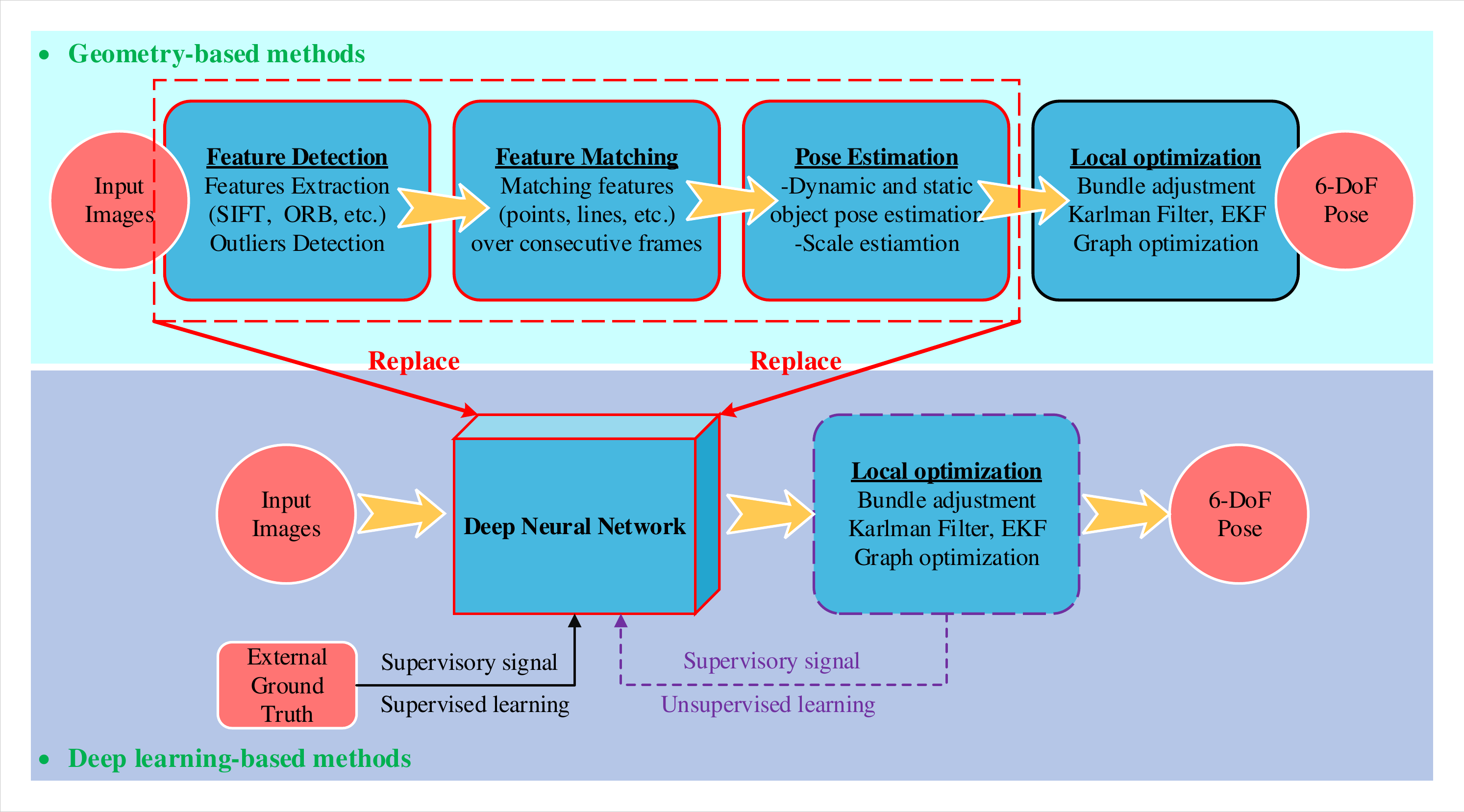}
\DeclareGraphicsExtensions.
\caption{Geometry-based and Deep learning-based visual odometry paradigms. The geometry-based visual odometry computes the camera pose from the image by extracting and matching feature points. The deep learning-based visual odometry can estimate the camera pose directly from the data. For supervised visual odometry, it requires external ground truth as supervision signal, which is usually expensive. In contrast, the unsupervised visual odometry uses its output as supervision signal. Besides, the local optimization module is optional for deep learning-based visual odometry.}
\label{figure1}
\end{figure*}

Recently, the development of mobile devices especially smart-phone make visual odometry more accessible for common users, which derivatives many emerging applications and also brings many challenges. Unfortunately, relying solely on classical geometry-based methods cannot meet these challenges, as these methods are highly dependent on the low-level hand-designed features that do not represent well the complex environment in the real world. Besides, the geometry-based method assume that the world is static. In practice, the scene geometry and appearance of the real-world changes significantly over time. Deep learning shows a powerful capability in many computer vision tasks, such as object detection, semantic segmentation, etc. This prompted the researcher to think about the possibility of applying deep learning to visual odometry. 

{\renewcommand\arraystretch{1.3}
\begin{table*}
\caption{\label{table1}Summarization of a number of related surveys since 2015}
\newcommand{\tabincell}[2]{\begin{tabular}{@{}#1@{}}#2\end{tabular}}
  \centering
  \begin{tabular}{cm{8cm}ccm{6cm}}
\toprule[2pt]
\textbf{No} & \tabincell{c}{\textbf{Survey Title}} & \tabincell{c}{\textbf{Ref}} & \tabincell{c}{\textbf{Year}} & \tabincell{c}{\textbf{Content}} \\
\midrule[1pt]
1 & \tabincell{l}{Visual Simultaneous Localization And Mapping: A Survey} & \tabincell{c}{\cite{RN108}}& \tabincell{c}{2015}&\tabincell{l}{Odometry\\Geometry-based Methods}  \\
2 & \tabincell{l}{Review of Visual Odometry: Types, Approaches, Challenges, \\and Applications} & \tabincell{c}{\cite{RN105}}& \tabincell{c}{2016}&\tabincell{l}{ Odometry \\Geometry-based methods}  \\
3 & \tabincell{l}{Past, Present, and Future of Simultaneous Localization and Mapping: \\Toward the Robust-Perception Age.} & \tabincell{c}{\cite{RN104}}& \tabincell{c}{2016}&\tabincell{l}{Metric and Semantic SLAM \\Open problems and future direction}  \\
4 & \tabincell{l}{A Critique of Current Developments in Simultaneous Localization \\And Mapping} & \tabincell{c}{\cite{RN151}}& \tabincell{c}{2016}&\tabincell{l}{Feature-based SLAM \\Pose graph SLAM using estimation and \\optimization techniques}  \\
5 & \tabincell{l}{Multiple-Robot Simultaneous Localization and Mapping: A Review.} & \tabincell{c}{\cite{RN103}}& \tabincell{c}{2016}&\tabincell{l}{SLAM  \\Miltiple Robot}  \\
6 & \tabincell{l}{Ongoing Evolution of Visual SLAM from Geometry to Deep \\Learning: Challenges and Opportunities.} & \tabincell{c}{\cite{RN58}}& \tabincell{c}{2018}&\tabincell{l}{ SLAM   \\Geometry-based approach  \\Deep learning-based approache}  \\
7 & \tabincell{l}{Visual SLAM and Structure from Motion in Dynamic Environments: \\A Survey.} & \tabincell{c}{\cite{RN61}}& \tabincell{c}{2018}&\tabincell{l}{SLAM  \\Structure from Motion  \\In dynamic environments}  \\
8 & \tabincell{l}{Simultaneous Localization and Mapping in the Epoch of Semantics: \\A Survey.} & \tabincell{c}{\cite{RN149}}& \tabincell{c}{2019}&\tabincell{l}{SLAM  \\Semantic concept}  \\
\bottomrule[2pt]
\end{tabular}
\end{table*}
}

Comparing with the classical geometry-based method, deep learning can automatically learn more effective and robust feature representations form large-scale datasets without the process of manual feature design. It seems that the interest point extracted by learning model is more robust to changing environment. Besides, by integrating semantic information, it becomes possible for the system to acquire a high-level understanding of the environment and achieve task-driven perception. Many advanced researches have shown that learning-based methods can achieve more robust performance under certain specific conditions, and have demonstrated the huge potential of integrated deep learning to solve the problems faced by geometry-based methods. Therefore, there is an unavoidable trend to use learning-based methods in visual odometry to improve the performance. Fig.\ref{figure1} shows the comparison of the geometry-based and deep learning-based visual odometry pipeline.

As shown in table \ref{table1}, there are some review papers about visual odometry or SLAM. Most recent review papers focus on geometry-based methods, less on deep learning-based approaches\cite{RN104, RN108, RN151}. Some review papers focus on specific aspects including multi-robot SLAM\cite{RN103}, learning-based SLAM\cite{RN58}, semantic information\cite{RN149},  and dynamic environment processing\cite{RN61}. However, these review papers not mainly focus on visual odometry. Aqel et al.\cite{RN105} was the only review that discussing visual odometry separately. But this review did not focus on the deep learning-based methods, and the deep learning-based method is a promising approach to improve the accuracy and robustness of visual odometry. 

In summary, we can find that most existing review papers focus on SLAM in the perspective of geometry. A review specific to visual odometry based on deep learning is still lacking. Therefore, in this paper, we provide a review on deep learning-based visual odometry including techniques, applications, open questions and opportunities. \textbf {In particular, our main contributions are as follows: }
\begin{itemize}
\item We propose a list of criteria as a uniform benchmark to evaluate the performance of visual odometry based on deep learning. 
\item We review the literature of visual odometry based on deep learning in three aspects by using the offered criteria as the uniform measurements. Then, we analyze and discuss the advantages and challenges of these methods in detail. Finally, how deep learning can improve the performance of visual odometry is also commented
\item We sum up the application of visual odometry and point out that Deep VO will play a big role in complicated and emerging areas such as search and rescue, planetary exploration, service robots. 
\item Based on the completed review and in-depth analysis, we put forward a unmber of open problems and raise some opportinities. 
\end{itemize}

The rest of this paper is organized as follows: In section \ref{section2}, a list of criteria is proposed, which serve as a uniform benchmark to evaluate the performance of the algorithms. In section \ref{section3}, we review the literature in three aspects based on the proposed criteria. In section \ref{section4}, we sum up the application of visual odometry. Challenges and opportunities are given in section \ref{section5}, followed by a conclusion in section \ref{section6}.
%----------------------------------------------------------------------------------------------------------------------
\section{Criteria of deep learning for visual odometry}
\label{section2}
In this section, we propose a number of criteria on deep learning-based visual odometry to evaluate the literature in a uniform measure manner in the next section. We identified 6 criteria that deep visual odometry algorithm should satisfy in order to obtain a more robust, accurate and high-level understanding, enabling a widespread application in complicated and emerging areas. \\
They are described below: 

 \textbf{(1) Accuracy}: refers to how accurate localization is. It is the most important criteria for evaluating learning-based visual odometry models or algorithms. We divide the literature into two categories: the algorithm with specific results and the algorithm without specific results. For the former, we use RMSE (Root Mean Square Error)\cite{RN224} to measure the bias of ground truth and estimation, which directly describes algorithm accuracy. Besides, for the literature without specific results, the different problem needs specific analysis. \\
Average rotational RMSE drift: 
\[{E_{rot}}({\cal F}) = \frac{1}{{|{\cal F}|}}\sum\limits_{(i,j) \in {\cal F}} \angle  \left[ {\left( {{{\widehat{\bf{p}}}_j} \ominus {{\widehat{\bf{p}}}_i}} \right) \ominus \left( {{{\bf{p}}_j} \ominus {{\bf{p}}_i}} \right)} \right]\]\\
Average translational RMSE drift:
\[{E_{trans}}({\cal F}) = \frac{1}{{|{\cal F}|}}\sum\limits_{(i,j) \in {\cal F}} {{{\left\| {\left( {{{\widehat{\bf{p}}}_j} \ominus {{\widehat{\bf{p}}}_i}} \right) \ominus \left( {{{\bf{p}}_j} \ominus {{\bf{p}}_i}} \right)} \right\|}_2}} \]

Where ${\rm {\mathcal F}}$ is a set of frames $(i, j)$, $\hat{\mathbf{p}} \in S E(3)$ and $\mathbf{p} \in S E(3)$ are estimated and true 
camera poses respectively, $\ominus$ denotes the inverse compositional operator and  $\angle[\cdot]$ is the rotation angle.

The notations used to evaluate these methods are introduced below. 

High (H): the algorithm has a lower RMSE drift in both translation and rotation, and provides enough experimental results and analysis to demonstrate. 

Medial (M): the algorithm has lower RMSE drift in translation or rotation, and provides not enough experimental results and analysis to demonstrate.
 
Low (L): the algorithm has large RMSE drift in translation and rotation, and provides not enough experimental results and analysis to demonstrate.

 \textbf{(2) Efficiency}: refers to how the algorithm use resources economically and effectively. For different application scenarios, system resources are limited in many aspects, such as computation, system memory. Using as less as possible resources to process as more as possible data in as shorter as possible time is a universal goal for learning-based visual odometry algorithms. We use computation time to reflect the algorithm efficiency to demonstrate algorithm advance through comparison with other algorithms. 

The notations used to evaluate these methods are introduced below. 

Yes (Y): since we use computation time as major criteria, Y means the computation time can reach real-time and uses resources economically and effectively. 

No (N): means the computation time not reach real-time and not uses resources economically and effectively. 

 \textbf{(3) Scalability}: refers to the ability of the algorithm to treat large-scale or long-term environments. For autonomous vehicles or other robots that works in large-scale outdoor environments, the algorithm should be able to work in continuous time with finite system memory. Further, the algorithm should find sufficiently discriminative features to against the changes (season, weather, trees leaves, etc.) in long-term environments. 

The notations used to evaluate these methods are introduced below. 

Yes (Y): the algorithm can handle well in large-scale or long-term environments, which is supported by experimental results. 

No (N): the algorithm cannot handle well in large-scale or long-term environments or this property is not corned in the paper. 

 \textbf{(4) Dynamicity}: refers to how the algorithm is able to deal with dynamic objects in environments. Visual odometry techniques usually based on the assumption that the observed environment is static. The dynamic object can cause the wrong pose estimation. Therefore, alleviating the impact of dynamic objects is an effective but challenging approach  to improve performance.  

The notations used to evaluate these methods are introduced below. 

Yes (Y): the algorithm can alleviate the influence of dynamic objects in environments. 

No (N): the algorithm cannot reduce the impact of dynamic objects in environments or this property is not corned in the paper. 

 \textbf{(5) Practicability}: refers to whether the algorithm can be used in practical applications. To deploy an algorithm, it usually considers many aspects such as efficiency, accuracy, and extensibility. It includes the trade-off between efficiency and accuracy. It also includes easy further improved and widely used.  

The notations used to evaluate these methods are introduced below. 

Yes (Y): the algorithm can be used in practical applications with a balanced trade-off between efficiency and accuracy, and tested in real-world dataset or practice. 

No (N): the algorithm is not suitable to applied in practical applications. 

 \textbf{(6) Extensibility}: refers to how algorithms or models can be easily further improved and widely used in various situations. First, the algorithm can automatically update the latest or temporary changes, and easy to maintain. Finally, the algorithm or model trained in a specific environment that can be generalized well to `unseen' environments without fine-tuning. 
\begin{figure*}[!t]
\centering
\includegraphics[width=7in]{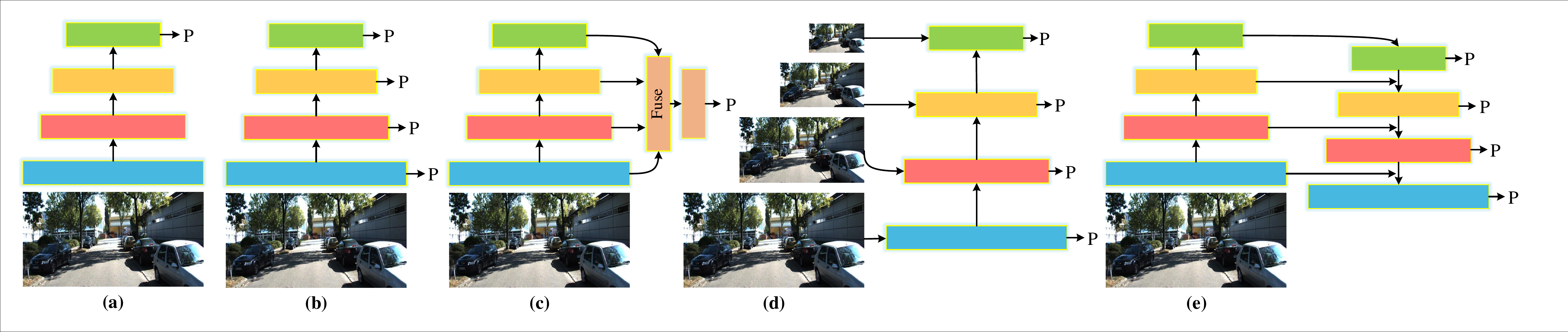}
\DeclareGraphicsExtensions.
\caption{An illustration and comparison of the paradigm for multi-scale feature learning used in depth estimation. (a) No multi-scale feature learning method is employed. (b) Depth is predicted on each feature map. (c) Depth is predicted on a single feature map, which is generated from multiple feature maps. (d) Depth is predicted from different scale images. (e) Before predicting depth on each feature map, the intermediate features from different scales are combined.}
\label{figure2}
\end{figure*}
The notations used to evaluate these methods are introduced below.

Yes (Y): the algorithm can be easily further improved and widely used in many situations. 

No (N): the algorithm cannot be improved or this property is not corned in this paper. 

\section{Deep learning for visual odometry} 
\label{section3}
\subsection {Deep learning for depth estimation}
Direct methods often require depth information to estimate the robot’s pose. And depth estimation is the beginning of the combination of deep learning and visual odometry. These methods can be divided into two categories: supervised and unsupervised learning methods. 
\subsubsection{Supervised learning methods}
For supervised learning, the use of supervised-based methods was first proposed in \cite{RN212}. They proposed a multi-scale neural network for depth estimation from a single image. The network architecture consists of two components: one for coarse global predictions and one for refined local predictions. Since then, a lot of work has utilized the multi-scale features of the network and depth structure information to improve the performance of depth estimation \cite{RN141, RN201, RN131, RN241, RN206, RN123}. Fig.\ref{figure2} is an illustration and comparison of the paradigm for multi-scale feature learning in depth estimation. 

For depth estimation problem definition, previous work usually makes depth estimation as a regression problem. Therefore, Liu et al.\cite{RN148} proposed a continuous CRF and CNN based framework, which transforms the depth estimation into learning problems. Besides, depth estimation can also be treated as a classification problem\cite{RN140}. The intuition behind this is that it is easier to estimate the depth range than to estimate a specific value. 

As far as we know, using CNN alone cannot model the long-range context well. Therefore, Grigorev et al. \cite{RN144} proposed a hybrid network by combining convolutional layer and ReNet layer. The ReNet layer consists of Long Short-Term Memory units (LSTMs), so the ReNet layer can obtain a global context feature representation. Moreover, an LSTM-based architecture was also proposed to explore whether RCNN can learn accurate spatial-temporally depth estimation without inter-frame geometric consistency or pose supervision\cite{RN184}. 

To obtain more accurate and robust results, the researchers turned their attention to more challenging and specific problems. To alleviate the depth ambiguity problem, invariant surface normal was used to assist depth estimation\cite{RN125}. And the ambiguity between focal length and monocular depth was demonstrated in \cite{RN135}. Ma et al. \cite{RN188} used both additional sparse depth samples acquired with a low-resolution depth sensor or computed via SLAM algorithms and input images to estimate the full-resolution depth. Additionally, a novel progressive hard-mining network was proposed in \cite{RN124} to solve the problem of large semantic gap and accumulated error over scales. DFineNet was proposed in \cite{RN232} to refine the depth estimation under sparse and noisy conditions. 

\subsubsection{Unsupervised learning methods }
Although the supervised learning-based method has achieved promising results in depth estimation tasks, the major problem faced by supervised learning is that it requires explicit ground-truth information that is difficult to collect in some scenarios. Moreover, the collected dataset unavoidable exist errors because of the sensor accuracy limitation.
\begin{figure*}[!t]
\centering
\includegraphics[width=6.5in]{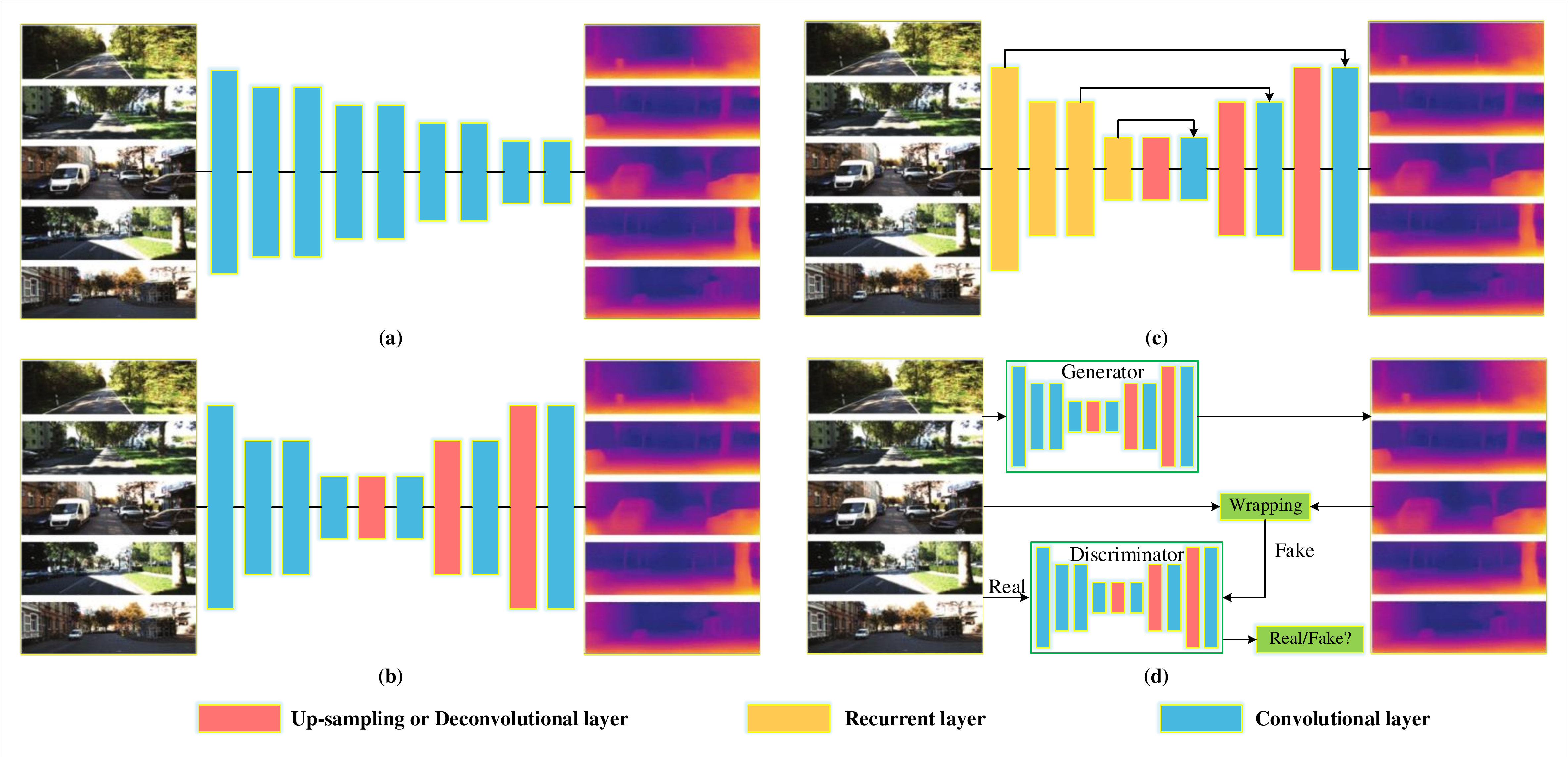}
\DeclareGraphicsExtensions.
\caption{An illustration of different neural network architectures for depth estimation. (a) CNN-based framework, which cannot obtain full resolution depth map. (b) CNN + Up-sampling or De-CNN based framework, which can obtain full resolution depth map. (c) RCNN based framework, which can obtain forward-backward consistent depth map. (d) GAN based framework. }
\label{figure3}
\end{figure*}
The method using unsupervised learning was first introduced in \cite{RN210}, which estimates the depth information from single view images. This framework is analogous to an autoencoder, in which the convolutional encoder takes the left image to predict the inverse depth map, and the decoder reconstructs the left image by synthesizing the right image with the predicted depth map. Similarly, Godard et al. \cite{RN200} introduced a method that can simultaneously infer disparities that warp the left images to match the right one or warp the right image to match the left one, thereby obtaining better depth estimation by enforcing left-right depth consistency. Zhou et al. \cite{RN197} first synthesized a new image of the scene from different views of the input image. They then synthesized the target image with per-pixel depth information by adding pose and visibility in consecutive views. Inspired by this work, Wang et al.\cite{RN127} utilized the camera parameters to synthesize the original view in a differential way. 

Different from the way of synthesizing the target view, GAN (Generative Adversarial Network) is more straightforward. MonoGAN \cite{RN229} first applied GAN to monocular depth estimation. The intuition behind this is that the generator utilizes the image to obtain the depth map and then uses the depth map to warp the image. Next, the discriminator was used to classify the raw image as real and the warped image as fake. Compare with MonoGAN, the generator in \cite{RN202} consists of DepthNet and PoseNet, which are used for depth map estimation and camera pose estimation, respectively, and uses the estimated depth map and camera pose parameters to generate image pairs. Fig.3 shows the different neural network architectures used for depth estimation.

Although using deep learning alone can achieve good results, it is still a problem how to bring geometry information or semantic information to depth estimation to improve the performance. Epipolar constraints play an important in geometry-based methods. Thus, epipolar constraints were used in learning-based methods to ensure proper pixel correspondence \cite{RN228} and reduce the network parameters \cite{RN395}. Similar to geometry-based methods using stereo image sequence to obtain absolute scale depth, learning-based methods also used stereo image sequences to train the network to solve the scale ambiguity problems \cite{RN328, RN66}. To enforce consistency between the left and right disparities, bilateral cyclic consistency constraints were used in \cite{RN231}. 2D photometric and 3D geometric information was also used in \cite{RN203}, which can consider the global scene and its geometric information. Besides, optical flow\cite{RN405} and semantic information was also used into depth estimation problem\cite{RN230}.

{\renewcommand\arraystretch{1.3}
\begin{table*}[htbp]
\caption{\label{table2}Summary and comparison of deep learning techniques for depth estimation. }
\centering
\begin{tabular}{ccccccccccc}
\toprule[2pt]
 \textbf{Ref} &  \textbf{Method Type} &  \textbf{Sensor} &  \textbf{Environment Structure }&  \textbf{Open Source} & \textbf{ Ac} & \textbf{ Ef} & \textbf{ Sc} &  \textbf{Dy }&  \textbf{Pr} &  \textbf{Ex} \\\midrule[1pt]
\cite{RN148} & Supervised & Mo & In/Outdoor & Y & L & N & A & A & N & Y \\
\cite{RN140} & Supervised & Mo & In/Outdoor & N & H & A & A & A & A & N \\
\cite{RN123} & Supervised & Mo & In/Outdoor & N & H & N & A & A & N & N \\
\cite{RN132} & Supervised & Mo & In/Outdoor & N & L & A & A & A & N & N \\
\cite{RN131} & Supervised & Mo & In/Outdoor & N & H & A & A & A & A & N \\
\cite{RN241} & Supervised & Mo & Indoor & Y & M & A & A & A & A & Y \\
\cite{RN125} & Supervised & Mo & In/Outdoor & N & M & N & A & A & N & N \\
\cite{RN188} & Supervised & Mo & In/Outdoor & Y & L & A & A & A & N & Y \\
\cite{RN124} & Supervised & Mo & In/Outdoor & N & M & N & A & A & N & N \\
\cite{RN707} & Supervised & St & Indoor & N & H & A & A & A & A & N \\
\cite{RN210} & Unsupervised & Mo/St & Outdoor & N & M & A & A & A & N & N \\
\cite{RN200} & Unsupervised & Mo/St & Outdoor & Y & H & A & A & A & Y & Y \\
\cite{RN197} & Unsupervised & Mo & Outdoor & Y & M & Y & A & A & Y & Y \\
\cite{RN404} & Unsupervised & Mo & Outdoor & N & L & A & A & A & N & N \\
\cite{RN66} & Unsupervised & Mo/St & Outdoor & Y & M & A & A & A & Y & Y \\
\cite{RN328} & Unsupervised & Mo & Outdoor & Y & L & Y & A & A & N & Y \\
\cite{RN163} & Unsupervised & Mo & Outdoor & Y & H & N & A & A & Y & Y \\
\cite{RN335} & Unsupervised & Mo & Outdoor & Y & M & Y & A & A & Y & Y \\
\cite{RN229}& Unsupervised & Mo & Outdoor & Y & H & A & A & A & A & Y \\
\cite{RN394} & Unsupervised & Mo & Outdoor & N & M & A & A & A & N & N \\
\cite{RN230} & Unsupervised & Mo & Outdoor & Y & L & A & A & A & N & Y \\
\cite{RN235} & Unsupervised & Mo & Outdoor & N & H & Y & A & A & Y & N \\
\cite{RN203} & Unsupervised & Mo & Outdoor & Y & M & Y & A & A & Y & Y \\
\cite{RN231} & Unsupervised & Mo & Outdoor & Y & M & A & A & A & A & Y \\
\cite{RN234} & Unsupervised & Mo/St & Outdoor & N & H & N & A & A & N & N \\
\cite{RN709} & Unsupervised & Mo/St & Outdoor & Y & H & A & A & A & Y & Y \\
\cite{RN697} & Unsupervised & Mo & In/Outdoor & N & H & A & A & A & A & N \\
\cite{RN710} & Unsupervised & Mo & Outdoor & N & H & A & A & A & A & N \\
\cite{RN673} & Unsupervised & Mo & Outdoor & N & H & Y & A & Y & Y & N \\
\midrule[1pt]
\multicolumn{11}{l}{\multirow{3}{*}{\begin{tabular}[c]{@{}l@{}} \textbf{Sensor: Mo} (Monocular)  \textbf{St} (Stereo)\\ \textbf{Ac: }Accuracy  \textbf{Ef: }Efficiency  \textbf{Sc:} Scalability  \textbf{Dy:} Dynamicity  \textbf{Pr:} Practicability  \textbf{Ex:} Extensibility \\  \textbf{Ac:} H (High) M (Medial) L (Low)  \textbf{Ef }\&  \textbf{Sc} \&  \textbf{Dy }\&  \textbf{Pr }\& \textbf{ Ex: }Y (Yes) N (No) A (Absence or No mention)\end{tabular}}} \\
\multicolumn{11}{l}{} \\
\multicolumn{11}{l}{} \\ \bottomrule[2pt]
%Sensor: Mo (Monocular) St (Stereo) &  &  &  &  &  &  &  &  &  &  \\
%Ac: Accuracy Ef: Efficiency Sc: Scalability Dy: Dynamicity Pr: Practicability Ex: Extensibility &  &  &  &  &  &  &  &  &  &  \\
%Ac: H (High) M (Medial) L (Low) &  &  &  &  &  &  &  &  &  &  \\
%Ef or Sc or Dy or Pr or Ex: Y (Yes) N (No) A (Absence or No mention) &  &  &  &  &  &  &  &  &  & 
\end{tabular}
\end{table*}
}
\subsubsection{Discussion }
An overview of main approaches discussed in this section and whether they meet our criteria is described in table \ref{table2}. Although binocular cameras can obtain depth information by utilizing stereo matching algorithms, it needs enormous computation and cannot handle well texture-less environments. Monocular cameras have the characteristics of low cost and its application is becoming more popular. For VO or SLAM, obtaining a coarse depth information is a great improvement to the convergence and robustness. Therefore, obtaining depth information directly from monocular images has received great attention. 

Based on table \ref{table2}, we can see that almost all approaches can predict depth under outdoor environments, which is the basis for the system to work in complex and large-scale environments. However, most approaches cannot meet the criteria of efficiency because of deep model usually needs enormous computation, which cannot run real-time in resources limited systems. Further, it hard to apply to real-world applications. Besides, the supervised method relying solely on the dataset has poor generalization performance, making it difficult to obtain a model that is common to most scenarios. Therefore, unsupervised methods attract more attention. For example, some work uses stereo image sequences or the relationship between camera pose and depth to obtain supervision signals. Besides, there are also some attempts to estimate depth information by adding sparse depth information obtained by other simpler sensors or from a sparse depth map. 

From this short analysis, it seems that the problem of building an accurate, effective and practical algorithm or model remains a challenge. Regarding accuracy, even though some impressive results are shown, it is still difficult to deploy in real-world applications according to the aspect of practicality. Besides, it is also an important problem to explore what features the neural networks have learned in the problem of depth estimation. 

\subsection{Deep learning for feature extraction and matching}
\subsubsection{Feature extraction }
Feature points are usually composed of key points and descriptors. Finding and matching then across images has been attracted large attention of researchers. The descriptor can be used to capture important and distinctive pixel points in an image, which plays an important role in feature extraction and matching task. To learn consistent descriptors, a novel mixed-context loss and scale-aware sampling method were proposed in \cite{RN268}. Mixed-context loss takes advantage of the scale consistency of Siamese loss and the fast learning ability of triplet loss. To learn compact binary descriptors, an unsupervised-based method was introduced in \cite{RN255}, which enforces the criterion of minimal loss quantization, uniformly distributed code and uncorrelated bits. Besides, Yi et al. \cite{RN244} proposed a unified framework for local feature detection and descriptors, and Zeng et al. \cite{RN248} used a 3D convolutional layer-based framework to learn local geometric descriptors. 

From another perspective, previous work used deep layer’s features, but shallower layer’s features are more suitable for matching tasks. Therefore, HiLM \cite{RN259} combined the different layers’ features to learn more effective descriptors. Because deep learning needs numerous data in the network training process, it usually a time-costing process. Progressive sampling strategies\cite{RN271, RN279} were used to allow the network to traverse a large amount of training data in a short time. Besides, to improve the descriptor geometric invariance and discrimination, a subspace pooling\cite{RN276} was proposed to instead of the max-pooling or average pooling. 

Since we don't know which points are "interesting", we can't find true "interesting" points by using manually labeled data. Unsupervised methods can automatically find “interesting” points from the data. Thus, Savinov et al. \cite{RN252} introduced a method that can generate repeated interesting points even if the image undergoes a transformation. Then, map the detected points to real-values and rank the points according to the real-values. Finally, the top/bottom quantiles of these points are used as interesting points. However, the scoring network response curve is constrained when the training process relies only on simple ranking losses. Therefore, a loss function was developed in \cite{RN278} that can maximize the peak value of the response map and improve the repeatability of the learner. 
\begin{figure*}[!t]
\centering
\includegraphics[width=6.5in]{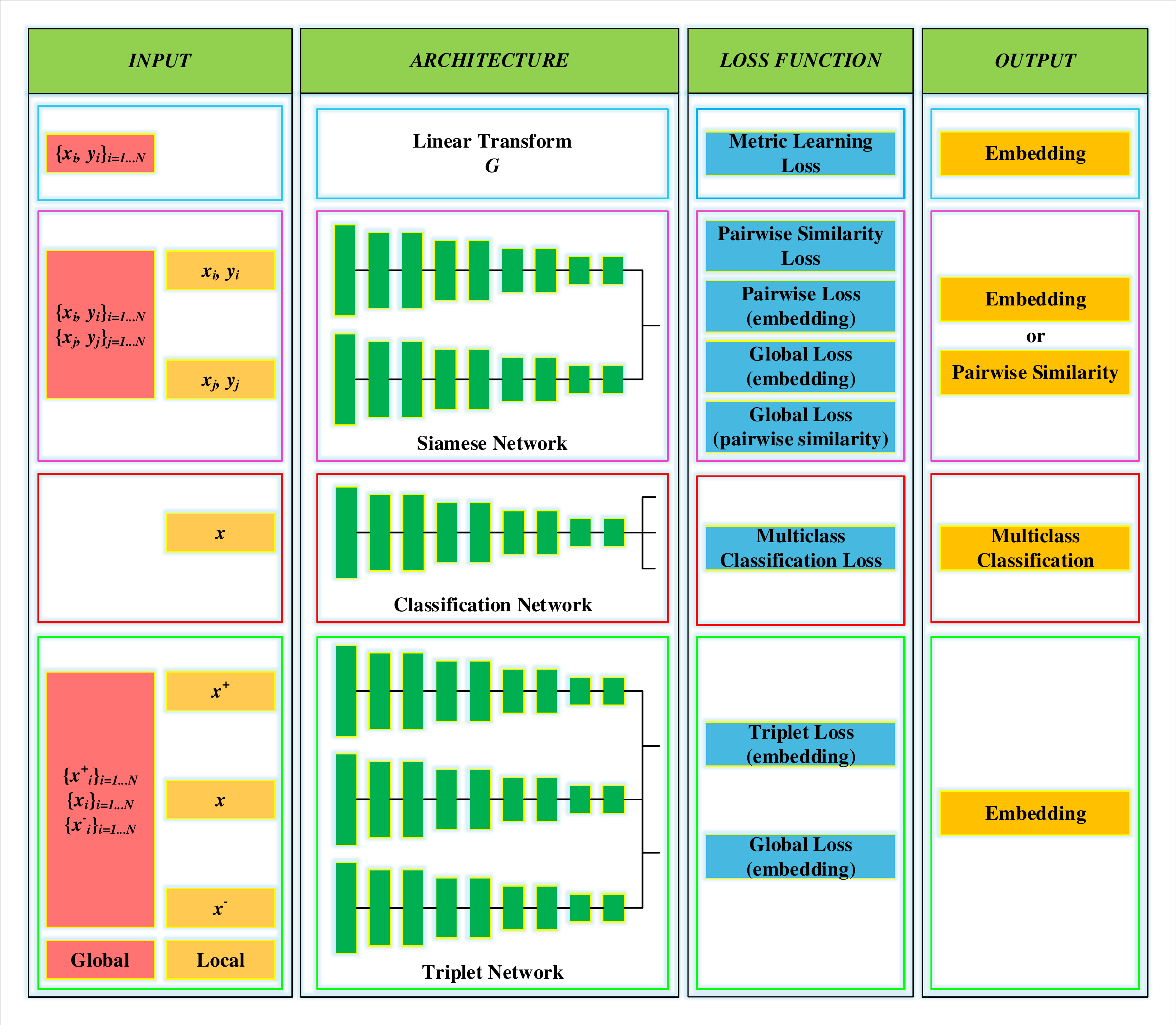}
\DeclareGraphicsExtensions.
\caption{Comparison of different types of input, network architectures, loss functions and output used to train the method of local image descriptor models\cite{RN269}. From top to bottom: (a) Linear Transform G trained by the metric learning loss and produce a feature embedding. (b) Different types of loss functions and input types can be used to train the Siamese network and produce a feature embedding (x) or a pairwise similarity estimation (xi, xj). (c) The Classification Network can be used to classify local image descriptors and used for multiclass classification problems. (d) The Triplet Network can also be trained using different kinds of loss functions and input types and produces a feature embedding. x+ represents a point from different class of x, x and x- represents a point from a different class of x. }
\label{figure4}
\end{figure*}
\subsubsection{Feature matching }
For feature matching, most previous work focused on Siamese and Triplet loss functions. Gadot et al. \cite{RN247} used Siamese CNN to learn the descriptors of both images, and then compared the learned descriptors using the L2 norm. The core of this method is a novel loss function, which can compute each training batch’s higher loss distribution moments. Additionally, a global loss function was proposed in \cite{RN269} that can minimize the mean distance between the same class descriptors and maximize the mean distance between the different class descriptors. Then, this work was extended in \cite{RN250} by proposing a regularization term to spread out the local feature descriptor in descriptor space, and the proposed regularization can improve all methods using pairwise or triplet loss. 

Contrast to traditional work that matches points according to the descriptors. Cieslewski et al. \cite{RN263} proposed a method for matching interesting points without descriptors. The proposed network has multiple output channels so that the corresponding points of two images can be matched implicitly by the channel ids. For example, if the i-th interesting point is the maximum of the i-th channel, that point should match the i-th channel in another image. Because there are no matching descriptors, the system requires less memory and less computational cost. Although this method cannot achieve the performance of traditional methods, it can generate confidence for specific interesting points. 

Further, a CNN- and RNN-based framework was proposed in \cite{RN257}. The importance of this work is that the network does not learn some specific descriptors, but learns how to find the descriptor that can be matched well.
 
\begin{figure*}[!t]
\centering
\includegraphics[width=6.5in]{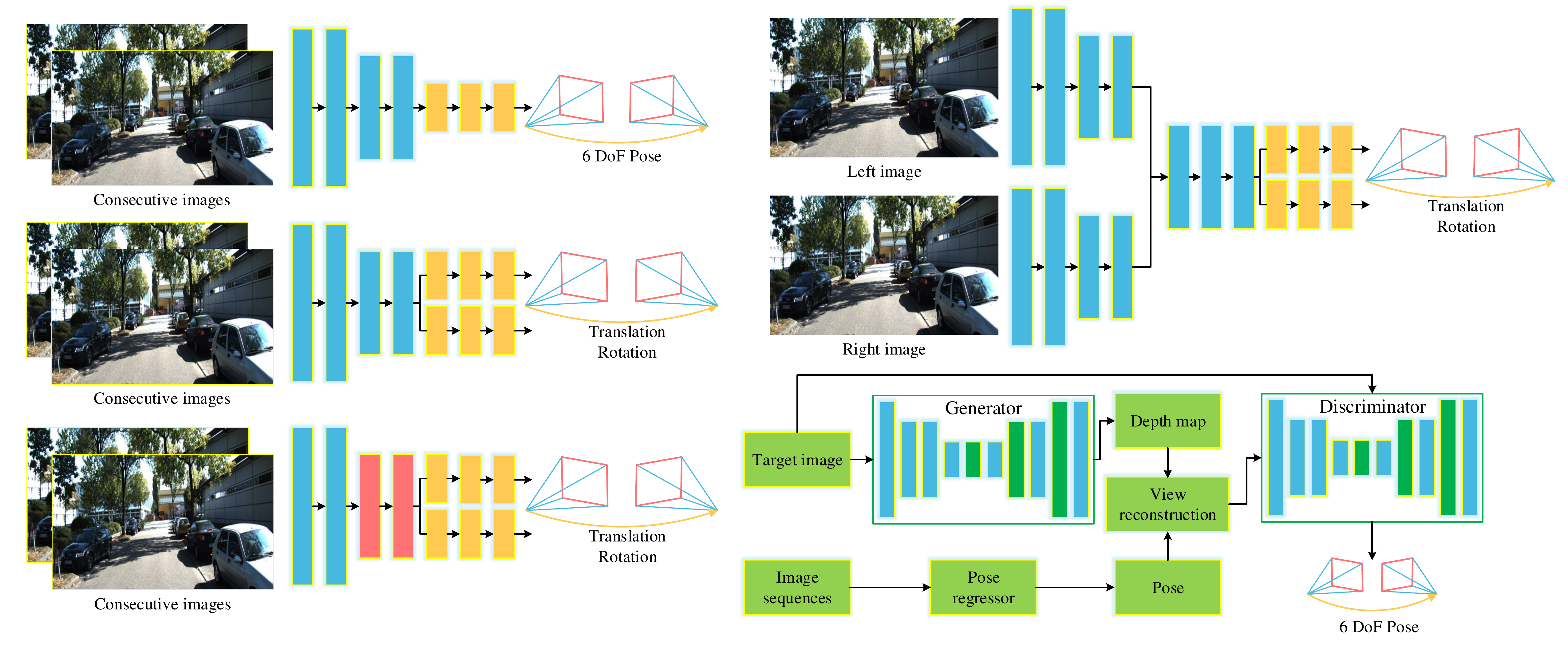}
\DeclareGraphicsExtensions.
\caption{ Description of different neural network architectures for pose estimation. (a) CNN-based framework, which produces a 6 DoF pose. (b) CNN-based framework with two fully connected networks to produce translation and rotation, respectively. (c) RNN-based framework, which produces a forward-backward consistent pose estimation. (d) Stereo based framework, which uses stereo image to obtain absolute scale pose estimation. (e) GAN based framework.     }
\label{figure5}
\end{figure*}
\subsubsection{Discussion}
Due to the proposed criteria is more focus on the full visual odometry system, we don’t use these criteria to evaluate the literature about feature extraction and matching. But to a certain extent, learning-based methods can alleviate the algorithm’s dependence on manual features. By utilizing the learning capability of deep learning, algorithms can directly learn more effective and robust feature representations. However, the efficiency of learning-based methods still needs to be improved. And how to deploy it to systems with limited resources remains a challenge. With the popularity of mobile devices, fast and robust feature extraction and matching for mobile devices is also a trend for future development. On the other hand, the improvement of the accuracy of learning-based methods is based on large-scale data in the training process. Therefore, in some scenarios without sufficient training dataset, it is still difficult to use deep learning techniques. Besides, is there something missing from the popular feature learning methods such as Siamese and Triplet architectures? Finally, although CNN can extract the information contained in the patch and establish the invariance of complex geometry and illumination changing, is this learned invariance overly rely on the data and cannot be effectively generalized to the complex changes between the images in the real world?

\subsection{Deep learning for pose estimation}
\subsubsection{Supervised methods}
PoseNet \cite{RN354} was the work that the first to utilize CNN to regress 6 DOF camera pose from the monocular image in an end-to-end manner without graph optimization. In large baselines that SIFT-based methods fail sharply, the proposed method can handle well. Besides, this method is also robust to illumination changing, motion blur and different camera intrinsics. Then, this work was extended in \cite{RN353} by utilizing multi-view geometry as a source of training data to improve the PoseNet performance. 

Since utilizing CNN to estimate camera pose from monocular images, it cannot model the relationship of consecutive images. Therefore, CNN and LSTM based pose regressor was introduced in \cite{RN374}, where CNN is used to learn suitable and robust feature representations, and LSTM is used to choose the most useful feature correlation for pose estimation. A spatial-temporal model based on bidirectional RNN was proposed in \cite{RN350} to learn the relationship between consecutive images. MagicVO \cite{RN398} used bidirectional LSTM to learn dynamic dependencies in consecutive images. In addition, DeepVO \cite{RN403} used RNN to model the relationship of motion dynamic and image sequences. ESP-VO \cite{RN375} then extended this work by adding uncertainty estimation for pose estimation. 

Due to the visual odometry will pass the error from the previous time to the next time, the error will accumulate over time, and drift is inevitable. To reduce the drift, an RCNN based framework DGRNet was proposed in \cite{RN390}. In this framework, the pose estimation sub-network was used to smooth the visual odometry trajectory and the pose regression sub-network was used to reduce the accumulation of camera pose errors. Compared with this method using only image streams, Peretroukin et al. \cite{RN366} used Bayesian Convolutional Neural Network to generate sun direction information in the image and then incorporated this sun direction information into stereo visual odometry pipeline to reduce drift. 

For the environment that geometry-based methods hard to handle, learning-based methods show a strong capability. Costante et al. \cite{RN388} proposed three different CNN architectures to learn robust representations to overcome the problems of blur, luminance and contrast anomalies. VLocNet \cite{RN325} learn separate discriminative features that can be well generalized to motion blur and perceptual aliasing environments.
{\renewcommand\arraystretch{1.3} 
\begin{table*}[htbp]
\caption{\label{table3}Summary and comparison of deep learning techniques for pose estimation. }
\centering
\begin{tabular}{ccccccccccc}
\toprule[2pt]
 \textbf{Ref }&  \textbf{Method Type }&  \textbf{Sensor }&  \textbf{Environment Structure }&  \textbf{Open Source} & \textbf{ Ac} &  \textbf{Ef }&  \textbf{Sc} &  \textbf{Dy} & \textbf{ Pr }&  \textbf{Ex} \\\midrule[1pt]
\cite{RN354} & Supervised & Mo & In/Outdoor & Y & L & Y & Y & N & N & Y \\
\cite{RN353} & Supervised & Mo & In/Outdoor & N & M & Y & Y & N & N & N \\
\cite{RN374} & Supervised & Mo & In/Outdoor & N & M & A & Y & N & N & N \\
\cite{RN350} & Supervised & Mo & In/Outdoor & N & M & Y & Y & N & Y & N \\
\cite{RN398} & Supervised & Mo & In/Outdoor & N & H & A & Y & N & N & N \\
\cite{RN403} & Supervised & Mo & Outdoor & Y & L & N & Y & N & N & Y \\
\cite{RN375} & Supervised & Mo & In/Outdoor & N & L & N & Y & N & N & N \\
\cite{RN388} & Supervised & Mo & Outdoor & N & M & Y & Y & N & Y & N \\
\cite{RN406} & Supervised & St & Outdoor & Y & M & A & Y & N & N & Y \\
\cite{RN325} & Supervised & Mo & Outdoor & N & H & Y & Y & N & Y & N \\
\cite{RN236} & Supervised & Mo & Outdoor & N & M & A & Y & N & N & N \\
\cite{RN698} & Supervised & Mo & Outdoor & N & H & Y & Y & N & Y & N \\
\cite{RN703} & Supervised & St & In/Outdoor & N & H & N & Y & Y & Y & N \\
\cite{RN204} & Unsupervised & Mo & In/Outdoor & Y & M & A & Y & N & N & Y \\
\cite{RN203} & Unsupervised & Mo & Outdoor & Y & L & Y & Y & N & N & Y \\
\cite{RN376} & Unsupervised & Mo & Outdoor & N & L & A & Y & N & N & N \\
\cite{RN394} & Unsupervised & Mo & Outdoor & N & M & A & Y & N & N & N \\
\cite{RN167} & Unsupervised & Mo & Outdoor & N & M & Y & Y & Y & Y & N \\
\cite{RN335} & Unsupervised & Mo & Outdoor & Y & M & A & Y & Y & N & Y \\
\cite{RN337} & Unsupervised & Mo & Indoor & Y & M & Y & N & N & N & Y \\
\cite{RN66} & Unsupervised & Mo/St & Outdoor & Y & M & A & Y & N & N & Y \\
\cite{RN328} & Unsupervised & Mo/St & Outdoor & Y & H & Y & Y & N & Y & Y \\
\cite{RN163} & Unsupervised & Mo/St & Outdoor & Y & M & N & Y & N & N & Y \\
\cite{RN358} & Unsupervised & Mo & Outdoor & N & H & N & Y & N & N & N \\
\cite{RN396} & Unsupervised & Mo & Outdoor & Y & H & A & Y & Y & N & Y \\
\cite{RN389} & Unsupervised & Mo & Outdoor & N & M & A & Y & N & N & N \\
\cite{RN706} & Unsupervised & Mo & In/Outdoor & N & H & Y & Y & N & Y & Y \\\midrule[1pt]
\multicolumn{11}{l}{\multirow{3}{*}{\begin{tabular}[c]{@{}l@{}} \textbf{Sensor: Mo }(Monocular)  \textbf{St }(Stereo)\\ \textbf{Ac: }Accuracy  \textbf{Ef: }Efficiency \textbf{ Sc:} Scalability  \textbf{Dy:} Dynamicity  \textbf{Pr: }Practicability \textbf{ Ex:} Extensibility \\  \textbf{Ac:} H (High) M (Medial) L (Low)  \textbf{Ef} \&  \textbf{Sc} \& \textbf{ Dy }\&  \textbf{Pr} \&  \textbf{Ex: }Y (Yes) N (No) A (Absence or No mention)\end{tabular}}} \\
\multicolumn{11}{l}{} \\
\multicolumn{11}{l}{} \\ \bottomrule[2pt]
%Sensor: Mo (Monocular) St (Stereo) &  &  &  &  &  &  &  &  &  &  \\
%Ac: Accuracy Ef: Efficiency Sc: Scalability Dy: Dynamicity Pr: Practicability Ex: Extensibility &  &  &  &  &  &  &  &  &  &  \\
%Ac: H (High) M (Medial) L (Low) &  &  &  &  &  &  &  &  &  &  \\
%Ef or Sc or Dy or Pr or Ex: Y (Yes) N (No) A (Absence or No mention) &  &  &  &  &  &  &  &  &  & 
\end{tabular}
\end{table*}
}
\subsubsection{Unsupervised methods}
For unsupervised learning-based methods, most of them use depth and optical flow to assist the training process\cite{RN372, RN400, RN701, RN711}. DeMoN \cite{RN204} was the first to use unsupervised learning methods to estimate both depth and pose from consecutive images. This network used optical flow to assist the depth and motion estimation. SfM-Net \cite{RN400} simultaneously predict pixel-level depth, segmentation, and camera pose from consecutive images. D3VO\cite{RN702} incorporates depth estimation, pose estimation and pose uncertainty into a framework to improve the performance. Further, to learn consistent 3D structures and better exploit unsupervision, a forward-backward constraint based on the left-right consistency constraint \cite{RN200} was proposed. In contrast, Iyer et al. \cite{RN337} proposed Composite Transformation Constraints to generate supervisory signals during training and to enforce geometric consistency. 

The monocular visual odometry faced the scale ambiguity problem. Many methods used stereo images to train networks to remove this problem\cite{RN66,RN328,RN163}. Besides, different from using stereo images to obtain absolute scale estimation, using depth information obtained from 3D LIDARs to train the network to obtain absolute scale estimation was introduced in \cite{RN358}. 

Another problem that plagues VO is how to alleviate the impact of dynamic objects. To reduce the distraction of dynamic objects in the scene, Barnes et al. \cite{RN167} introduced an ephemerality mask that can estimate the likelihood that pixels in the input image correspond to static or dynamic objects in the scene. Additionally, GeoNet was proposed in \cite{RN335}, which consists of two stages and three sub-networks. The first stage is a rigid structure re-constructor for depth and pose estimation, and the last stage is a non-rigid motion localizer for optical flow estimation. Specifically, three subnetworks extract geometric relationships separately, and then combine as an image reconstruction loss to infer the static and dynamic scene parts in two stages, respectively. 

Since geometry-based methods becoming mature, how to utilize this geometry knowledge to improve learning-based methods is also a problem. To consider ambiguous pixels and obtain better geometric understanding ability, Prasad et al.\cite{RN228} proposed an epipolar geometry-based approach, which gives each pixel a weight according to whether it is projected correctly. Shen et al. \cite{RN396} proposed to incorporate intermediate geometric information such as pairwise matching to the pose estimation problem to solve the problem of results involving large systematic errors under realistic scenarios. VLocNet++ \cite{RN389} embedded geometric and semantic knowledge of the scene into the pose regression network. 

\subsubsection{Discussion}
An overview of the main approaches discussed in this section and how they whether meets our criteria is described in table \ref{table3}. As shown in table \ref{table3}, the current learning-based methods still not yet viable for robots working under dynamic environments. It remains a challenge for both learning-based and geometry-based methods. Another interesting observation is that even the accuracy is high that they also cannot be used in practice. The reason is that the learning-based model inference is a time-consuming process. Of course, things could evolve in the near future depending on the system computation power increasing. However, designing a lighter and small network with good results is also an evolving trend.
 
The main challenges for learning-based methods are how to improve efficiency and accuracy. For accuracy, some currently existing learning-based methods can compare with the state-of-the-art geometry-based methods. However, it builds on the similarity (refers to feature representation) between the training and testing dataset. If testing environments contain many ‘unseen’ scenarios, the performance will be decreased. However, a dataset contains all scenarios is impossible. For efficiency, most applications require to produce online estimation in a timely fashion. Unfortunately, they are resources limited and do not have GPU to accelerate inference. Last but not least, robustness. Due to the powerful learning capability of deep learning, it can improve the robustness to a certain extent. Nevertheless, this robustness is heavily dependent on large-scale data. We don’t know what the algorithm learned in this process. How to use existing excellent geometry-based algorithms to guide them to improve the robustness in challenging environments is a promising way. 
\section{Application of visual odometry in complicated and emerging areas}
\label{section4}
In this section, we will focus on visual odometry applications. Recently, visual odometry has a wide range of applications and has been effectively applied in several fields. 

In mobile robotic systems, the autonomous vehicle is a rapidly advancing application area of visual odometry with unstructured environments\cite{RN467, RN536, RN486}. Visual odometry is generally regarded as a key for truly autonomous vehicles. In this application, the algorithm more focus on dynamic \cite{RN167}, illumination changing \cite{RN489}, long-term \cite{RN186} and large-scale environments \cite{RN495}. Compared with autonomous vehicles, using visual odometry into UAVs is more challenges due to the computational capability and 3D maneuverability\cite{RN433, RN438}. In order to obtain more robust performance, many researches have tried to fusing data from multi-modal sensors, such as IMU\cite{RN549, RN550}. Recently, there has shown a growing interest in fusing multiple cameras\cite{RN473, RN472, RN435}. Besides, visual odometry is also widely used in underwater robots\cite{RN654, RN655, RN656, RN657, RN658, RN659}, space exploration robots\cite{RN662, RN661, RN660, RN663} and agriculture robots\cite{RN470}, etc. 

In the medical industry, VO has great potential in image-based medical applications, which can navigate outside or inside the patient’s body to refer position and discovered problems\cite{RN664}. For example, active wireless capsule endoscopes used VO techniques to track his location without additional sensors and hardware in the GI tract\cite{RN455, RN446, RN444, RN449, RN445, RN370, RN454}. Besides, in surgical applications, visual odometry also gained a big attention such as surgical systems, surgical assistance robot and image-guided surgery systems\cite{RN665, RN667, RN669, RN668}. 

\begin{figure}[!t]
\centering
\includegraphics[width=3.5in]{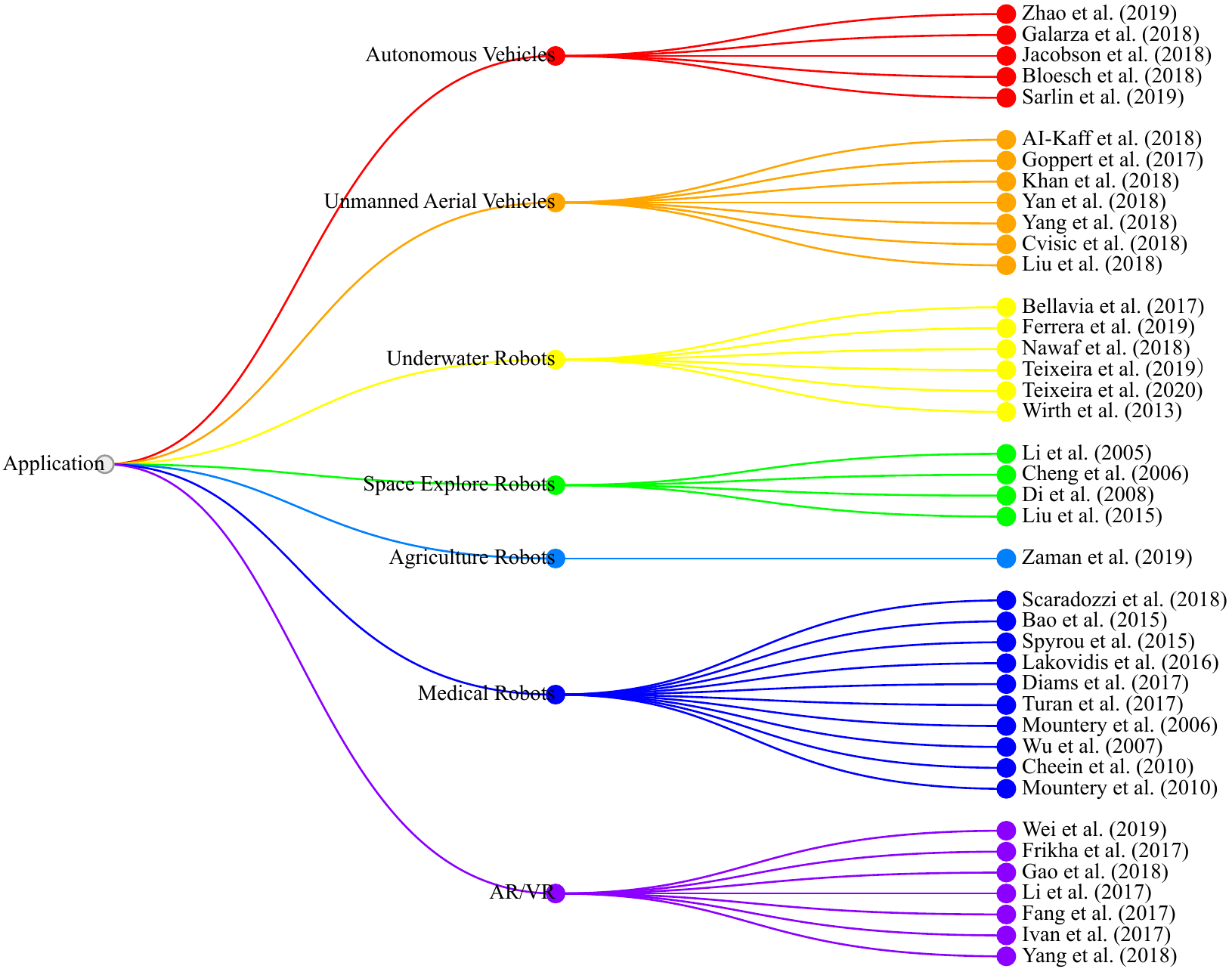}
\DeclareGraphicsExtensions.
\caption{Application of visual odometry. }
\label{figure6}
\end{figure}
In augmented reality (AR), VO also plays an important role. Visual odometry are used to obtain the device real world coordinates independent of camera and camera images\cite{RN524}. Then, integrate augmentations with the real world according to real world coordinates obtained by VO. In the beginning, these systems usually use camera as only sensors\cite{RN522, RN517}. In mobile devices, to improve accuracy and obtain absolute scale, the devices fuse data from other sensors such as IMU\cite{RN532, RN514, RN441,RN516}. 

Although visual odometry has a wide range of potential applications, the different application requires special hardware and software design. Each of these applications brings different challenges. It seems that deep learning is a promising way to handle these challenges. It has been demonstrated that deep learning can improve accuracy and robustness. Deep VO will play a big role in complicated and rising areas, including search and rescue, planetary exploration, service robots. 
\section{Challenges and opportunities}
\label{section5}
\subsection{Challenges}
\indent First, the data available to train the deep learning model for various tasks is insufficient. Although we know that large-scale high-quality data is required to train deep models, there is still a limited number of the large-scale dataset available in visual odometry or SLAM. The existing large-scale dataset like KITTI from the perspective of geometry-based methods is not large-scale from the perspective of deep learning. Besides, the scene of these existing datasets is single, which causes poor generalization ability of the model. The performance of deep learning-based methods can be improved as the amounts of datasets increase. Therefore, a large-scale and high-quality dataset is important to push learning-based methods, like ImageNet for object recognition and PASCAL VOC dataset for object detection. 

Second, deep learning methods used for visual odometry are simple. Since CNN surpasses human ability to classify object recognition in ImageNet 1000 challenge, CNN has almost dominated the field of computer vision. Therefore, CNN was first used to estimate the camera pose and has spawned a lot of work based on CNN. However, CNN is inadequate and less effective to learn relationships and dynamics from sequential data. Thus, RNN was used to learn these complex connections and dependencies in an end-to-end manner. Different methods have their own disadvantages and advantages. A problem of RNN is that they cannot extract effective feature representation from high-dimensional data like CNN. The power of deep learning should be far more than this. Therefore, bring new techniques and theories to visual odometry needs us to explore in the future. 

Third, we don’t know what features the neural networks have learned. For geometry-based methods, it is explicit and well understand. Although it is not robust to dynamic and illumination changing environments and needs plenty of time to manually fine-tune, we can know what the algorithm learned from this process. Nevertheless, for deep learning, we think more of it as a black box. We usually can only rely on the trained model to obtain the results and cannot predict theoretically. Even if the model results not meet our expectations, we do not know how to improve. 

Finally, the generalization ability and robustness of current algorithms need to be improved. Compared with geometry-based methods, learning-based algorithms usually don’t fail initialize and lose track. The reason is that given an image, the model always produces a prediction. But, it can make a big error due to wrong or “unseen” input. Besides, the similarity of feature representations between the training and testing dataset are the decisive factors affecting performance. Therefore, how to improve the generalization is a big problem. Based on table \ref{table3}, few literatures discussed the dynamic environments. Although the capability of deep learning improved the accuracy and robustness in specific situations, learning-based methods are still lacking an effective way to solve this problem. To improve the robustness of the algorithm is an important step to robust perception. 
\subsection{Opportunities}
(1) Use unsupervised learning techniques to train the network. Since deep learning models trained by supervised learning acquiring amount of human-labeled data, it cannot benefit from large-scale unlabeled data. The performance of unsupervised learning can be improved as the amounts of datasets increase. On the one hand, unsupervised learning can truly exploit high-dimension features from large-scale data. On the other hand, we can use the geometry-based loss function to guide the learning process in unsupervised learning. 

(2) Use semantic information to obtain semantic reasoning ability. Understanding semantic information is the most significant step to obtain high-level understanding and semantic reasoning. Currently, robots only understanding the low-level geometric features but they don’t truly exploit semantics. On the one hand, utilizing semantic information and object detection results can form semantic-level based localization constraints, thereby improving the accuracy and robustness. On the other hand, integrating semantic information into the visual odometry allows the robot to infer the surrounding environment. For example, when the system detects a vehicle, the system can infer the direction of the vehicle’s motion based on the direction and position of the vehicle. Moreover, understanding high-level semantic information, object properties and the properties' mutual relations can provide better interaction between the robot and the environment. 

(3) Try to fuse data from multi-modal sensors by using deep learning techniques. Since a single sensor is difficult to meet the requirement, many systems combined camera data with multi-modal sensor data by using filters or optimization techniques. However, few literatures focus on learning-based fusion methods. Compared with traditional fusion methods, learning-based methods do not require manual complex system modeling and calibration, such as synchronization and calibration between the camera and other sensors, modeling the sensor’s noise and biases, removing gravity from acceleration measurements according to orientation. Therefore, learning-based multi-sensor fusion methods have the potential to generate new adaptive VO paradigms that can be adapted to different sensors. 

(4) Combining deep learning-based methods with geometry-based methods. Geometry-based methods are more reliable and accurate under static, sufficient illumination and texture environments. However, they still face many problems in challenging environments. While learning-based methods can handle challenging environments, these methods cannot provide reliability and accuracy of geometry-based methods in conditions that suitable for geometry-based methods. Therefore, combining geometry-based methods with deep learning-based methods is an effective approach to improve system accuracy and robustness. However, it is not clear in which way this should be realized to be most effective. 

(5) Deploy learning-based algorithms in hardware platform efficiently. Although, with the development of microelectronics and mobile processing techniques, the computing power of embedded systems is still weak. Most applications usually use embedded systems as visual processing systems, and require to produce online estimation in a timely fashion. Nvidia’s Jetson TX1/2 or AGX Xavier can provide GPU for UAVs or mobile under low power costs and lightweight conditions. However, it doesn’t meet the requirements of most learning-based methods. Therefore, designing the lighter, smaller network with good enough results is a significant step to make these algorithms into practice. 
\section{Conclusion}
\label{section6}
This paper provides a review of the state-of-the-art for visual odometry based on deep learning, with main focus on literatures over the past five years. Various algorithms were reviewed, and applications were summed up. The advantages and disadvantages were also briefly discussed. We first propose a number of criteria to evaluate the algorithms reviewed in this paper. We further reviewed the literature in the aspect of depth estimation, feature extraction and matching, and pose estimation. Then, we sum up a wide range of applications of visual odometry and emphasize deep learning bringing an effective way to improve the robustness and high-level understanding. Finally, based on discussion and analysis, remains challenges and opportunities were presented.

\section*{Acknowledgment}
The authors thank the financial support of National Natural Science Foundation of China (Grant No: 51605054), Key Technical Innovation Projects of Chongqing Artificial Intelligent Technology (Grant No. cstc2017rgzn-zdyfX0039), Chongqing Social Science Planning Project (No:2018QNJJ16), Fundamental Research Funds for the Central Universities (No: 2019CDXYQC003).



\bibliographystyle{IEEEtran}


\bibliography{MyPaperReference}

\end{document}